\documentclass[journal ]{new-aiaa}
\usepackage[utf8]{inputenc}
\usepackage{textcomp}

\usepackage{graphicx}
\usepackage{amsmath}
\usepackage[version=4]{mhchem}
\usepackage{siunitx}
\usepackage{longtable,tabularx}
\usepackage{booktabs}
\title{Real-time processing of high-resolution video and 3D model-based tracking for remote towers}

\author{Oliver J.D. Barrowclough\footnote{Research Scientist, Dept. of Mathematics and Cybernetics, oliver.barrowclough@sintef.no.} 
and Sverre Briseid\footnote{Research Scientist, Dept. of Mathematics and Cybernetics, sverre.briseid@sintef.no.}
and Georg Muntingh\footnote{Research Scientist, Dept. of Mathematics and Cybernetics, georg.muntingh@sintef.no.}
and Torbj\o rn Viksand\footnote{Research Scientist, Dept. of Mathematics and Cybernetics, torbjorn.viksand@sintef.no.}
}
\affil{SINTEF Digital, Forskningsveien 1, 0373 Oslo, Norway}

\begin{document}

\maketitle

\begin{abstract}
High quality video data is a core component in emerging remote tower operations as it inherently contains a huge amount of information on which an air traffic controller can base decisions. Various digital technologies also have the potential to exploit this data to bring enhancements, including tracking ground movements by relating events in the video view to their positions in 3D space. The total resolution of remote tower setups with multiple cameras often exceeds 25 million RGB pixels and is captured at 30 frames per second or more. It is thus a challenge to efficiently process all the data in such a way as to provide relevant real-time enhancements to the controller. In this paper we discuss how a number of improvements can be implemented efficiently on a single workstation by decoupling processes and utilizing hardware for parallel computing. We also highlight how decoupling the processes in this way increases resilience of the software solution in the sense that failure of a single component does not impair the function of the other components. 
\end{abstract}




\section{Introduction}
Since the deployment of the first remote tower implementation in \"Ornsk\"oldsvik, Sweden in 2014 interest in remote tower has been growing internationally. Today there is a focus on validating the \emph{multiple remote tower} concept, in which an air traffic controller (ATCO) is responsible for more than one airport simultaneously \cite{papenfuss2016head,li2018much}. This movement of the concept towards more complex scenarios demands new technologies that support ATCO's situational awareness in order to ensure that their cognitive capacities are not exceeded. Such technologies may include providing visual enhancements or even fully automating parts of the ATCO's responsibilities. This requires processing of large amounts of data, including high-resolution video data, in real-time.  
Despite the ever-increasing capacity of modern computers, their computational power is still insufficient to allow for naive implementations when dealing with large amounts of data. There are, however, a number of solutions, both hardware and software based, that make such processing attainable. 

\subsection{Contribution of the paper}

In this paper we describe the implementation of several features that enhance remote tower based on raw video. In particular, we highlight the following contributions:

\begin{itemize}
    \item Introduce cost effective methods in the form of passive optical tracking of objects that do not rely on object movements or active surveillance hardware, such as radar and LIDAR. 
    \item Combine state-of-the-art object detection algorithms and suitable attention mechanisms to handle high throughput and extremely high resolution video data.
    \item Consolidate these (per frame) detections into (cross frame) tracked objects using combinatorial optimization.
    \item Develop and evaluate automatic correction of video exposure and white balance for a camera array, not requiring overlap between adjacent cameras and achieving real-time performance in our high-throughput setting.
    \item Position tracked objects in a minimap by combining a 3D aerodrome model with video tracking and present the minimap on the screens, enhancing heads-up time. 
    \item Develop a software architecture in which the various components utilize parallel processes on a single workstation. The software components communicate in such a way that failure of a non-vital component (e.g. tracking) does not affect the performance of the vital ones (i.e., video stream rendering).
\end{itemize}

\section{Background}

\subsection{Video processing}

Live streamed video is a core component of remote tower systems, including both light spectrum and infrared imaging. Given that the cameras are often placed some distance from the runway, high-resolution of the streamed imagery is considered an essential component. In contrast, it has been shown that video frame rate is of less importance when it comes to maintaining visual detection performance and does not impact physiological stress levels \cite{jakobieffects}. Nevertheless, it is desirable from the point-of-view of video quality and system perception to utilize as high a frame rate as the bandwidth allows. Remote tower data is most often transferred on high bandwidth networks meaning that, in many cases, high frame rates are available in addition to high resolution. Essentially, this means that a huge amount of information is being continually updated at a fast pace. 

\begin{figure*}
  \includegraphics[width=0.496\textwidth]{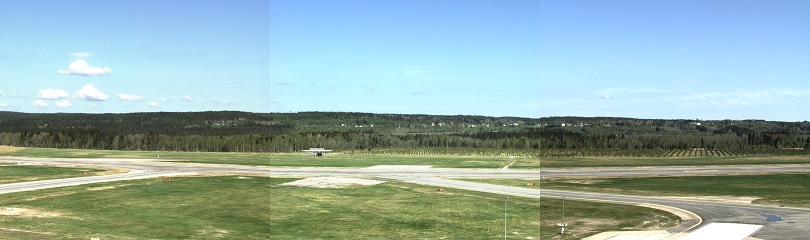}\hfill
  \includegraphics[width=0.496\textwidth]{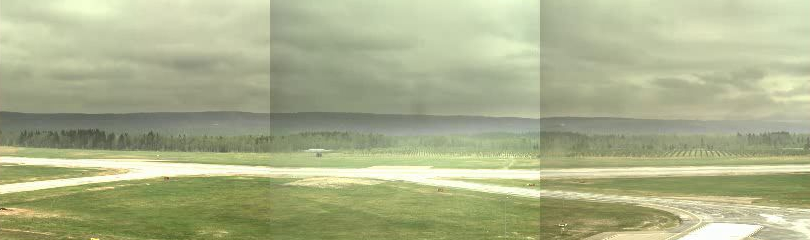}\vspace{0.5em} \\
  \includegraphics[width=0.496\textwidth]{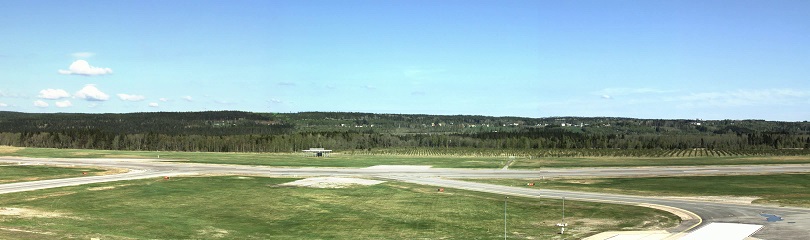}\hfill
  \includegraphics[width=0.496\textwidth]{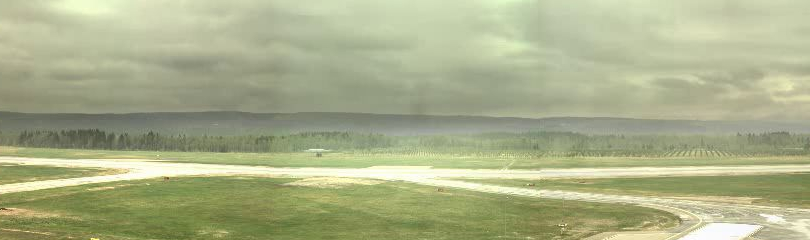}\\  \caption{Original (up) and exposure-corrected (down) images from the Saab remote tower at Sundsvall-Timr\aa~airport, for sunny (left) and cloudy (right) conditions. Images courtesy  LVF.}\label{fig:cams-original-corrected}
\end{figure*}

Another feature of remote tower implementations is that they often use multiple cameras to cover angles of up to 360 degrees. The exposure of each of these cameras is typically controlled individually in order to ensure that they present optimal contrast of the scene to the ATCO. For example, if the sun is shining directly towards one camera, the exposure profile required should be completely different to the profile of the camera pointing in the opposite direction. Nevertheless, this local correction of contrast often leads to visible `seams' between the images when presented side-by-side (see Figure \ref{fig:cams-original-corrected}). It is therefore of interest to apply local filters that smoothly adjust the contrast to avoid such seams appearing as prominently (stitching).

The aim of a stitching algorithm is to produce a visually plausible mosaic, in the sense that it is as similar as possible to the input images, but in which the seam between the stitched images is invisible \cite{levin:2004}.

There exist many methods for stitching panorama images and thus reducing the visible seam if the exposures do not match up. One example of such methods that has shown good results is Laplacian pyramid blending \cite{burt:1983} using a feathered blend. However, common to many of these methods in a panorama setting is that they operate on adjacent images with a region of overlap. In our scenario, we do not have such an overlap as the cameras are intended to be perfectly aligned. In \cite{pandey:2013} the authors describe stitching of non-overlapping images. They use a pyramid blending with a Savitzky-Golay filter to smooth the transition across the seam. The input images are unrelated and the result is a blurred area across the seam, which is visually pleasing but not what we aim for. Another method is gradient domain blending \cite{levin:2004}, where the method performs the operations in the gradient domain. The authors discuss two main image stitching methods. The optimal seam algorithm involves searching for a specific curve in the overlap region, thus not applicable in our scenario. The second method is relevant for our setup, using a minimization over the seam artifacts by smoothing the transition between the images.
It is, however, too compute-intensive to be run for every frame with our setup of up to 14 Full HD cameras.

\subsection{3D modelling for remote tower}

One major difference between remote towers and traditional towers is that the video presented in remote tower lacks any depth information. At the same time, accurate digital elevation models (DEMs) that capture 3D behaviour of terrain and other static objects such as buildings and trees are readily available. It is therefore of interest to explore methods for presenting 3D/depth information in the ATCO working position in order to enhance the situational awareness of ATCOs.

Although remote tower systems are often supplemented with pan-tilt-zoom cameras, the majority of cameras in a remote tower system are static. This gives the possibility to accurately calibrate the cameras with respect to a 3D coordinate system. Calibration involves determining the parameters of the cameras, that is: position, direction, tilt, field of view, aspect ratio and resolution. In some cases lens distortion is also an issue, but in our setting the problem is negligible. The advantage of calibrating the cameras is that any surface event that is detected on the video can be immediately positioned in 3D space \cite{barrowclough:2015}. A surface event is understood to be any event that is both visible on the video and that occurs on the surface of the airport, excluding only aerial targets. Such events could, for example, arise from an ATCO querying a certain location or from 2D video tracking software tracking the location of an object. 

A common format for DEM is GeoTIFF, which allows for encoding both a raster height map and georeferencing information in a single file. In our work we have had access to 2 m $\times$ 2 m resolution digital surface model (DSM) in GeoTIFF format that covered an area of approximately 30 km$^2$ in and around Sundsvall-Timr\aa{} airport. The vertical resolution of the data was 1 m.

\begin{figure}
    \centering
    \noindent
    \includegraphics[width=0.815\columnwidth,draft=false,clip]{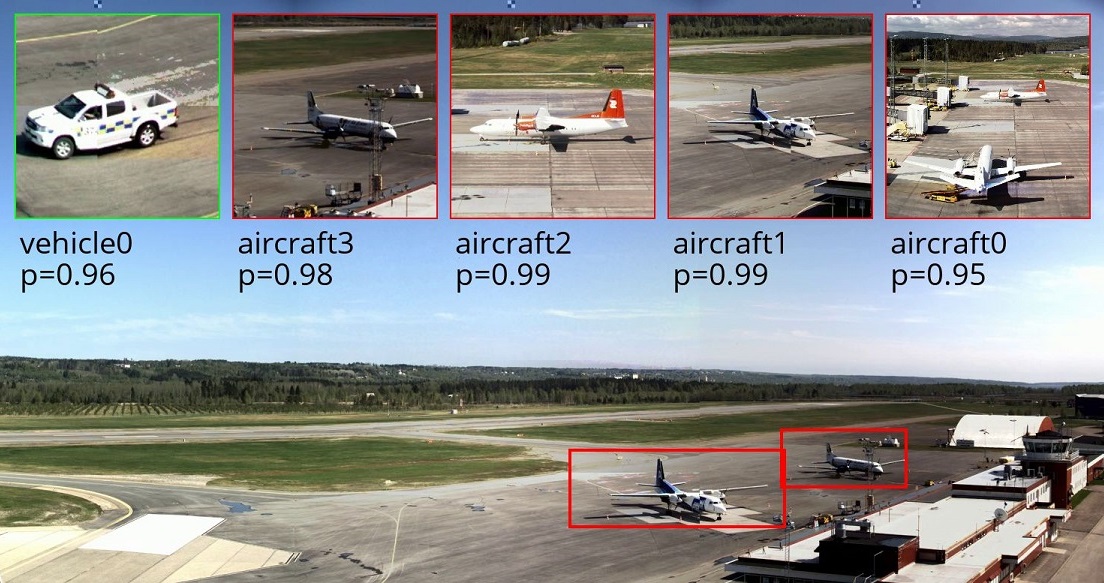}\vspace{0.2em}
    \includegraphics[width=0.2\columnwidth,draft=false,clip,trim=0 130 0 300 ]{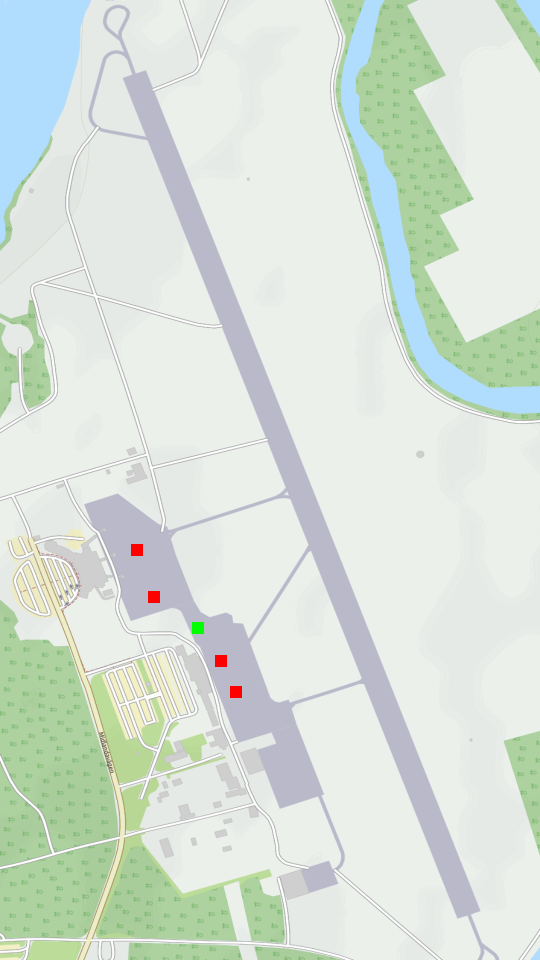} 
    \includegraphics[width=0.2\columnwidth,draft=false,clip,trim=0 130 0 300 ]{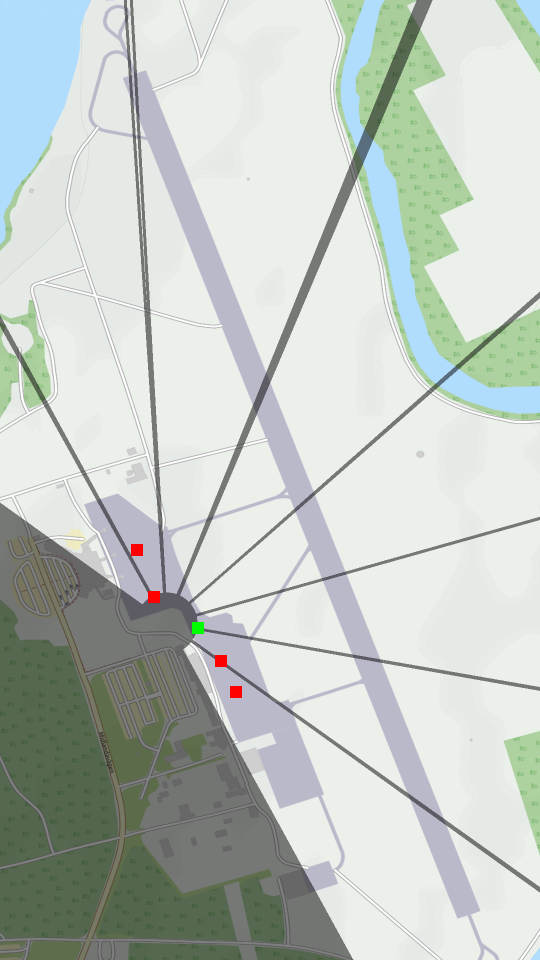} 
    \includegraphics[width=0.2\columnwidth,draft=false,clip,trim=0 130 0 300 ]{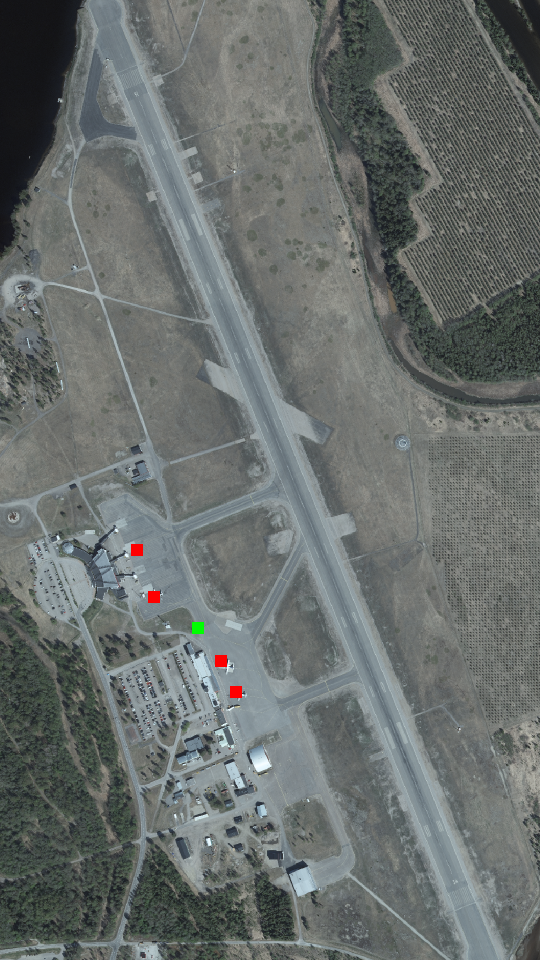} 
    \includegraphics[width=0.2\columnwidth,draft=false,clip,trim=0 130 0 300 ]{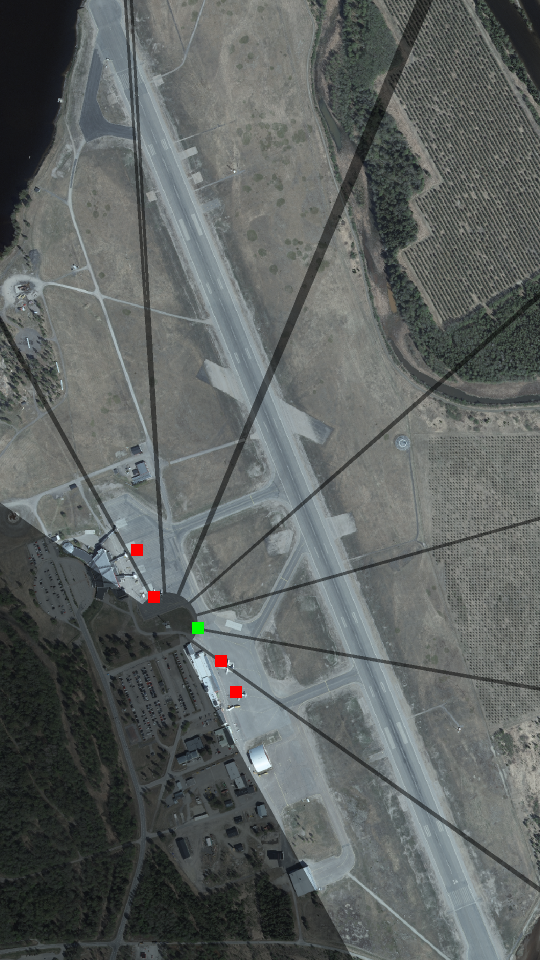}
    \caption{Tracked objects with close-up views and their position in maps overlays. Camera views courtesy LFV, maps \textcopyright\, OpenStreetMap CC BY-SA, orthophotos \textcopyright\, Lantmäteriet.}
    \label{fig:Map}
\end{figure}

\subsection{Tracking}
Tracking an object in a video scene is based on repeated detection and localization of the object on successive frames and obtaining a continuous association between detections through time. This association is made easier if a classification of the object is available. 

In recent years, machine learning algorithms have made great strides in performance, both with respect to computational efficiency and accuracy of the results. This has resulted in some tasks that were previously considered to be insurmountable, now exhibiting superhuman performance. In particular, this is true for:
\begin{itemize}
    \item \emph{detection} --- deciding whether an image contains an object;
    \item \emph{classification} --- determining the class of the dominating object in an image (in our case aircraft, vehicle, or person);
    \item \emph{localization} --- estimating the location of an object (in our case as a tight bounding box);
    \item \emph{tracking} --- locating a moving object over time.
\end{itemize}
In the context of remote tower, these tasks are of particular interest as they reflect some of the responsibilities of the ATCO. 

A popular method for real-time object detection from image data is known as YOLO (you only look once) \cite{redmon2016you,redmon2018yolov3}. YOLO frames the problem of detection and localization of objects in a scene as a regression problem that can be solved with a single evaluation of a neural network. This approach provides both better performance and is faster to evaluate than most of its predecessors. In addition to detecting and localizing objects, YOLO also provides a classification of the object and an estimated probability that the classification is correct.

Traditional approaches for real-time object tracking include methods such as mean shift \cite{comaniciu2000real}.
More recent approaches include making use of Siamese convolutional neural networks (CNNs) \cite{bertinetto2016fully}. In our setting, the main complication with respect to these approaches is that it is hard to achieve real-time tracking when considering the extremely high resolution of the data.

\subsection{Computational power}
Given one of the main drivers behind the remote tower concept is cost reduction, it is of interest to investigate how best to utilize computational resources. Despite the ever increasing performance of today's computers, particularly with respect to Graphics Processing Units (GPUs), the sheer amount of data to be processed in our setting raises a challenge. Nevertheless, we still consider it feasible to implement real-time video processing, tracking and 3D event localization on a single workstation equipped with one or more GPUs. Utilizing computational resources efficiently involves balancing the requirements of the different software components in terms of memory consumption (both main memory and GPU memory), and exploiting both multi-threading and highly parallel GPU processes such as matrix multiplication, which is a core-component of modern machine learning approaches. Tasks such as video decoding can be performed on fixed-function chips (e.g. NVIDIA's PureVideo) that are part of modern GPUs, enabling other GPU resources to be utilized for different processes.

\section{Methodology}
\subsection{Video processing}

\subsubsection*{White balance and exposure correction}
For a smooth transition between camera frames, narrow bands on each side of a seam should have close to identical colour spectra. Our fundamental assumption is the converse: If we are able to accurately match the colour spectra of adjacent narrow bands, then the transition will appear natural. We determined this assumption to be reasonable, as we expect the landscape to be approximately identical in these regions. This \emph{local constancy prior} holds as long as the video streams are well aligned geometrically and temporally. The video processing is performed for the synchronized frames.

\begin{figure}
    \centering
    \includegraphics[width=.3\columnwidth]{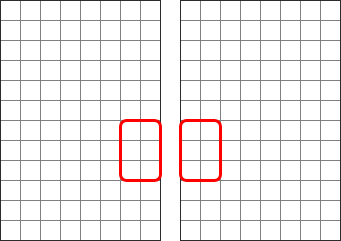}\quad
    \includegraphics[width=.3\columnwidth]{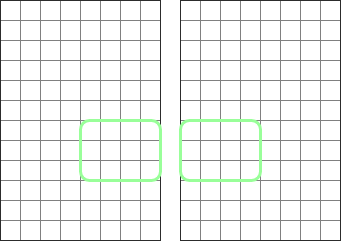}
    \caption{The exposure correction function is defined using a narrow band along the seam of adjacent images (red) and is applied to half the image (green).}
    \label{fig:ExpFuncDomain}
\end{figure}

To obtain a smooth transition, we estimate a shared spectrum at the seams based on averaging the intensity distributions in narrow bands along the seam (see Figure \ref{fig:ExpFuncDomain}). To flawlessly map from the measured spectra to the shared spectra would require a highly nonlinear and high-dimensional map. However, this would take too much time to compute, as well as being too slow to apply in real-time. Moreover, the measured spectra are only approximations of the underlying spectrum of the landscape, so a simpler mapping correctly matching the essential features of these distributions would be more appropriate and prevent overfitting.

GPUs are well known for their ability to swiftly apply linear operations. To keep our algorithm as efficient as possible, we consider an approach that only depends on adjacent video streams. Considering the 14 cameras in our setup, such a local approach yields a significant reduction in complexity. A further reduction is achieved by matching the spectra using a linear affine map $f_c(x) = a_c x + b_c$ for each individual colour channel $c = r,g,b$. These maps were applied from the middle of the adjacent images towards the common border (see Figure \ref{fig:ExpFuncDomain}). A gradual transition was obtained using convex combinations, from applying the identity map to the center of the image to the map $f_c$ on the border.

It turned out these maps were not expressive enough to accurately transform all dominant features in the spectra. For instance, in certain cases we obtained a map yielding a seamless transition in the sky but a poor transition on the ground (and vice versa). To overcome this issue, we decided to partition the stream vertically in blocks of identical size. 

Another issue, manifesting itself as a local flickering, appeared when an object moved from one stream to the next. In this case the pixels of this object suddenly outshine the pixels of the background landscape, dominating the colour spectrum and violating our fundamental assumption. To resolve this issue we implemented two supplemental methods.

The first method detects the movement using the thresholded absolute differences between frames (see Figure \ref{fig:ThresholdedDifference}), removing the corresponding pixels from consideration in the measured spectra when defining our exposure correction map. This \emph{object removal} approach is viable for objects that do not dominate the domain of the local exposure correction map. 

Should a moving object cover most or the whole block, there will be few or no pixels left for defining our map. For these cases we use an \emph{exponential smoothing} approach, which reduces the contribution of the moving object by blending the newly computed exposure function with the exposure function from the previous frame
\[ E_{\mathrm{blend}} := (1 - \alpha) \cdot E_{\mathrm{prev}} + \alpha \cdot E_\mathrm{new},\qquad \alpha = 0.05.\]

\subsection{3D modelling for remote tower}

The first step in combining the ``out the window'' video stream view with the 3D model is to calibrate the cameras in a 3D coordinate system. In our case we use the SWEREF99 geodetic reference system \cite{jivall2000sweref}. 

\begin{figure*}
    \centering
    \includegraphics[width=0.95\textwidth, clip, trim=0 0 0 330, draft=false]{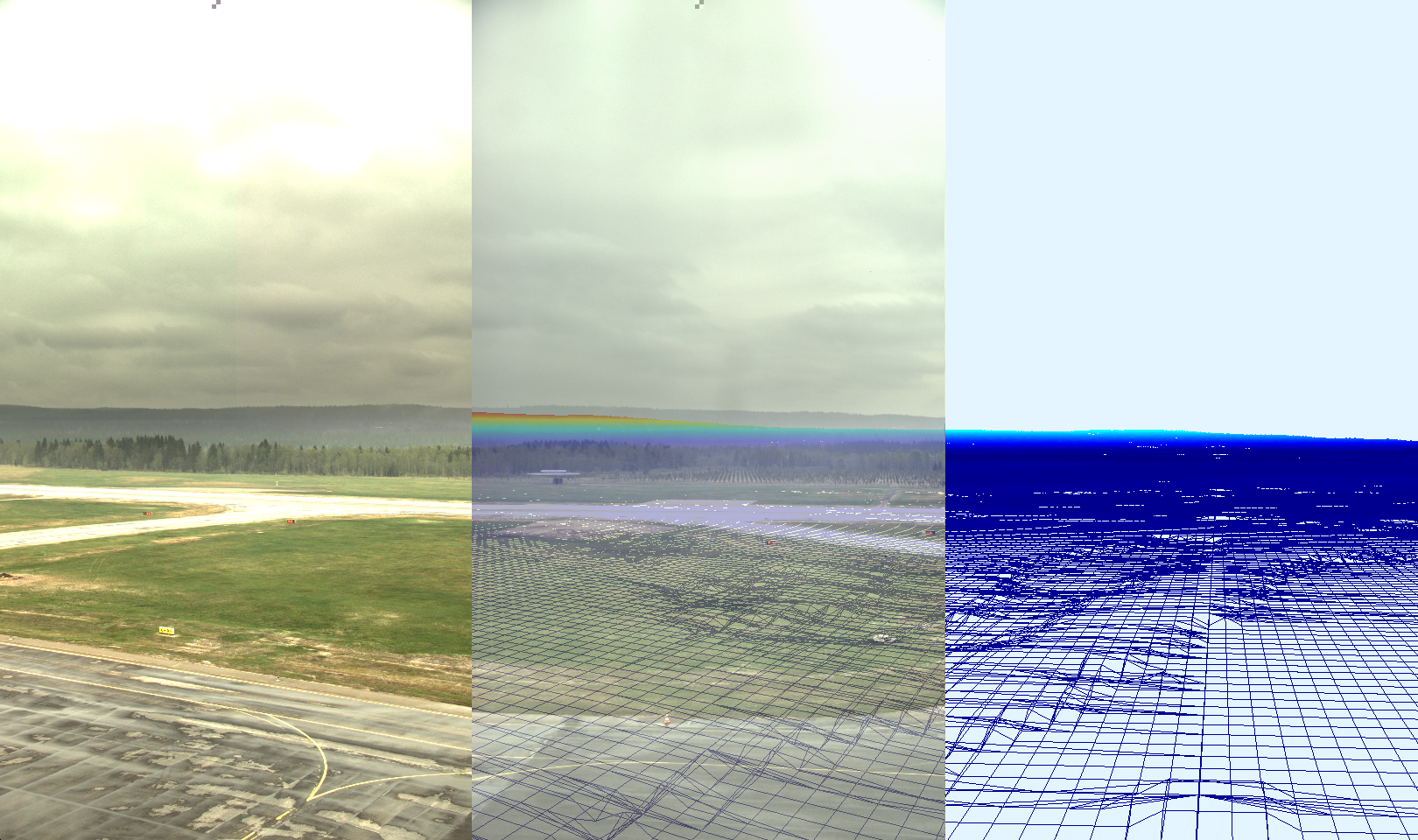}
    \caption{The video stream (left), 3D model (right), and blend (middle) in the camera calibration application.  Imagery courtesy LFV.}
    \label{fig:application}
\end{figure*}

To accurately calibrate the camera, it is normally possible to fix certain parameters in advance. For example, the position of the camera can normally be accurately determined by referring to map data, orthophotos or by GPS or equivalent positioning technologies, together with technical drawings of the remote tower structure. In addition, it is almost always the case that the cameras are horizontally aligned; that is, there is negligible tilt on the cameras. Information about the aspect ratio and resolution can easily be extracted from image metadata. The cameras considered in this paper also have negligible lens distortion, so non-linear correction is not required. Thus, it remains to define the direction and the field of view of the cameras. To aid this process, an interactive application has been implemented that allows both navigation in the 3D model and blending the video dynamically. The idea of the interactive application is depicted in Figure \ref{fig:application}. If that figure the horizon does not match up due to missing terrain data at large distances. The slight inconsistencies in the 3D model on the tarmac are due to rounding errors caused by the 1 m vertical resolution of the data. The known parameters can be supplied via a graphical user interface. The 3D navigation controls then allow the user to interactively align the features in the 3D model with the video imagery, and the parameters are updated dynamically. 

Based on the calibrated camera parameters, we can accurately compute the distance in 3D space between the camera origin and any point on the ground corresponding to a pixel location. These depths can either be computed by intersecting a ray with the 3D model \cite{barrowclough:2015}, or by simply rendering the scene with the depth buffer active. The resulting depths can then be used to position events in 3D space. They can either be computed on the fly or can be pre-computed for all pixels and stored in a depth map. In the spirit of decoupling the components as much as possible, we opt to pre-compute the depths. This results in a modest increase in memory requirements, but enables much better utilization of computational resources, freeing up CPU or GPU for other tasks.

\subsection{AI-based video tracking}

\subsubsection*{Detection}
In this work we use YOLO for detection, localization and classification \cite{redmon2016you,redmon2018yolov3}. YOLO is a convolutional neural network that extracts and uses the same features for classification and localization, in the form of multiple bounding box prediction. This makes the method both extremely fast and accurate, due to better generalization to unseen images achieved by this multitask learning.

Each detection consists of an object category (aircraft, vehicle, or person), axis-aligned bounding box, as well as a probability signifying the confidence of the detection. The three object classes were consistently colour-coded in their appearance as bounding boxes and close-ups in the video streams and as markers on the map (see Figure \ref{fig:Map}). To avoid false positives, only detections whose probability exceeds a threshold $tol_{\mathrm{detect}} = 0.65$ are processed.

The multiple detections returned by the YOLO architecture could correspond to the same object. Such superfluous detections are eliminated using greedy \emph{non-maximum suppression} \cite{Rothe2014NonmaximumSF} as follows: for each category, pick the detection with highest probability, and suppress overlapping detections within this category by setting their probabilities to zero. This process is then repeated for the remaining detections, until only detections with zero probability remain. 

Overlapping of bounding boxes $B, B'$ is quantified in terms of their \emph{Interection over Union} (IoU), defined as the quotient of the areas of their intersection and union, i.e.,
\[ \mathrm{IoU}(B,B') := \frac{\text{area}(B\cap B')}{\text{area}(B\cup B')}. \]
It measures similarity of the boxes, taking value 0 for disjoint boxes, value 1 for identical boxes, and otherwise values in between. Overlapping detections are then suppressed whenever their IoU exceeds a threshold $tol_{\mathrm{NMS}} = 0.45$.

However, YOLO cannot be applied directly to our situation, as it applies to square input images of fixed size. The image obtained by concatenating $n$ video streams is not square; it has size $1920\times (n\cdot 1080)$. Moreover, our high-end consumer grade GPU (GTX 1080 Ti, with 11Gb) runs out of memory, even when attempting to run YOLO on a $1600\times 1600$ subimage.

While memory is not an issue for running YOLO on an image of size $960\times 960$, the entire visual range can only be scanned once every couple of seconds in this manner. 
For the high spatial and temporal resolution of our setup, it is therefore important to develop effective \emph{attention mechanisms}, i.e., strategies for deciding where to look. 

\subsubsection*{Attention mechanisms}

\begin{figure}
    \centering
    \includegraphics[width=0.2\columnwidth]{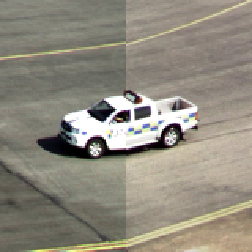}
    \includegraphics[width=0.2\columnwidth]{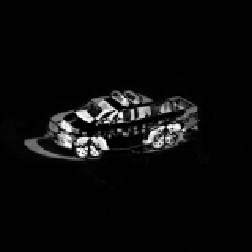}
    \includegraphics[width=0.2\columnwidth]{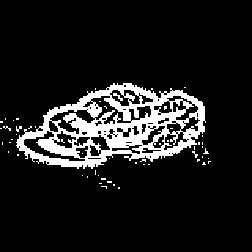}
    \caption{In a given image region (left), changes are detected by taking the absolute difference of consecutive frames (middle) and thresholding (right).}
    \label{fig:ThresholdedDifference}
\end{figure}

We consider the following three mechanisms:
\begin{enumerate}
    \item \emph{Sliding window approach}. After concatenating the frames of all 14 cameras in a single image, slide a fixed-sized window across this image and run a detection in each window. As an option, it is possible to use overlapping windows to avoid unfortunate cropping of objects. Another option is to (in addition) resize the image to detect objects at various scales.\\ This strategy is computationally expensive, and therefore only run on start-up to get a good overview of the initial situation.
    \item \emph{Difference approach}. Moving objects can be detected by detecting significant local changes in the video streams. Technically, this is achieved by thresholding the absolute difference of two consecutive frames, as shown in Figure \ref{fig:ThresholdedDifference}. Sliding a window across the resulting binary image, one runs a detection whenever the number of on-pixels (representing a significant change) exceeds a given threshold.
    \item \emph{Expectation approach}. Once we have an inventory of tracked objects with their locations and movements, we can predict its expected position in a future frame, and run a detection there.
\end{enumerate}

These mechanisms are combined in a high-level scheduler to effectively track objects, subject to the cost constraints imposed by the available computational resources.

\subsubsection*{Tracking algorithm}

Upon start-up of the tracker, one first applies the sliding window approach to yield an initial list of detections. During the remainder of the tracking process, the difference and expectation approaches are used for deciding where to run detections. Besides being used within each YOLO detection, non-maximum suppression is used here to remove superfluous detections by the various attention mechanisms. To avoid the creation of duplicate objects (and an ensuing cascade effect),
a low suppression tolerance $tol_{\mathrm{NMS}} = 0.3$ is used here.

The problem of optimally assigning a set of $m$ detections $D = \{d_i\}_i$ to $n$ existing objects $O = \{o_i\}_i$ can be expressed as an \emph{assignment problem}. For this, one first defines a cost function $C: D \times O\longrightarrow \mathbb{R}$, in which a higher cost reflects a less desirable match. The values of this function are assembled in a \emph{cost matrix}
\[
\mathbf{C} =
\begin{bmatrix}
C(d_1, o_1) & \cdots & C(d_1, o_n)\\
\vdots & \ddots & \vdots \\
C(d_m, o_1) & \cdots & C(d_m, o_n)
\end{bmatrix} \in \mathbb{R}^{m,n}.
\]
For the \emph{linear sum assignment problem}, the goal is to find a one-to-one assignment $f:D\longrightarrow O$, for which the total cost
\[
\sum_{i=1}^n C(d_i, f(d_i))
\]
is minimal. This problem can be solved rapidly (in cubic running time) using the Hungarian algorithm \cite{Kuhn1955}. Such an assignment problem is solved for every category separately.

Let $B_d$ and $B_o$ be the bounding boxes of detection $d$ and object $o$ measured at frame numbers $f_d$ and $f_o$.
To impose a penalty for \emph{dissimilarity}, we consider a cost function complementary to the IoU, defined by 
\[ C(d,o) := 1 - \mathrm{IoU}(B_d, B_o) \cdot a^{f_d - f_o},\qquad a = 0.99.\]
This function imposes a higher cost for matching a detection $d$ with an object $o$ last observed in a distant frame, by discounting their IoU by a factor $a$ for every frame that has since passed.

After finding the optimal assignment $f$, each detection $d$ is added to the history of the object $o = f(d)$ if
\[ 1 - C(d,o) > tol_{\text{IoU}} := 0.05, \]
i.e., if the discounted IoU exceeds a given tolerance. If this is not the case, as well as for the unassigned detections, it is checked whether 
\[ 1 - \min_{i=1} C(d, o_i) < tol'_{\text{IoU}} := 0.001, \]
i.e., whether the detection wasn't just outmatched, but not relevant for any of the existing objects. If this is the case, it is added as a new object. This rather strict tolerance avoids the duplication of objects due to inaccurate detections.

\begin{figure*}
    \centering
    \includegraphics[width=0.95\textwidth]{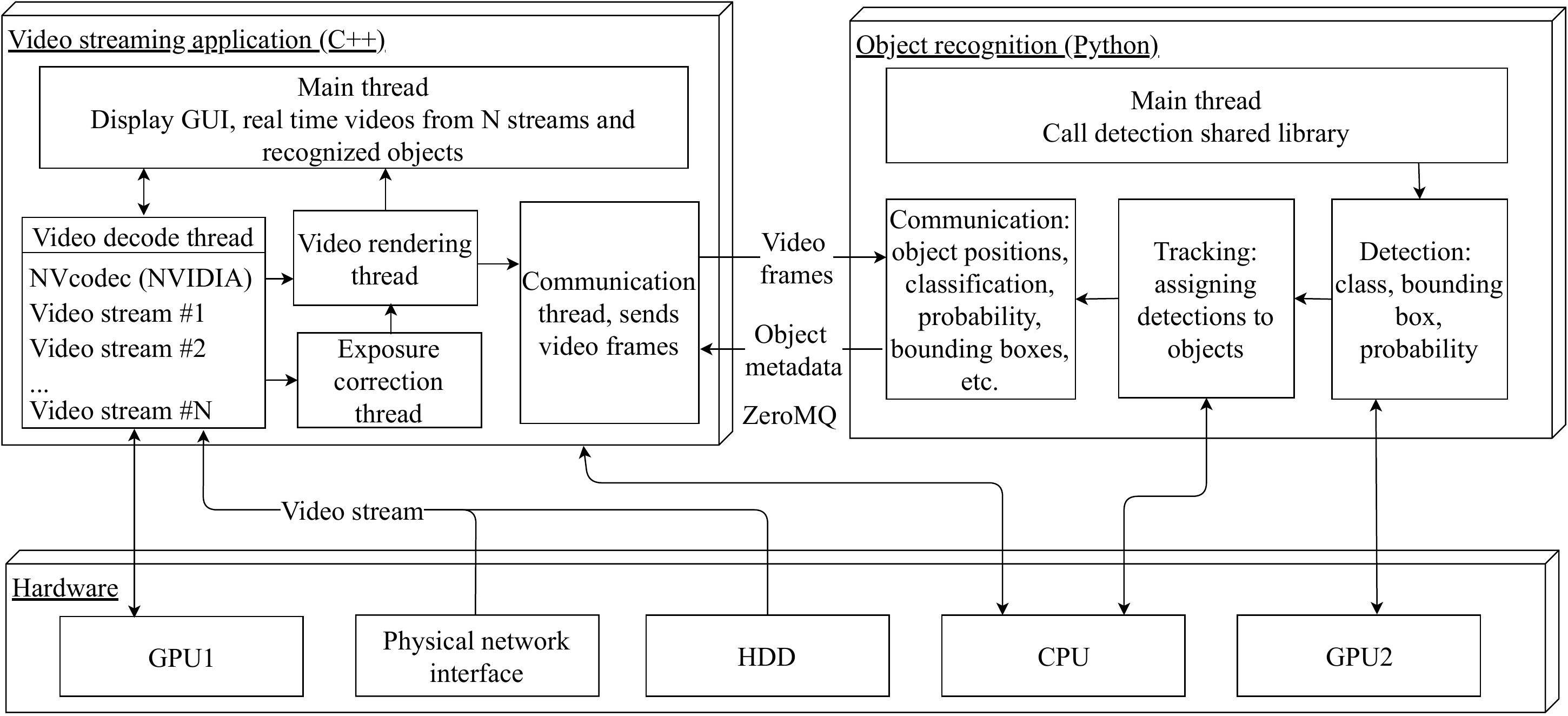}
    \caption{Application block diagram and mapping of the software processes to different hardware components} 
    \label{fig:block_diagram}
\end{figure*}

\subsection{System}
The video streams enter the system as H.264 compressed video streams \cite{Richardson:2010:HAV:1942939} in $1920\times 1080$ resolution. In our case, 13 such streams had to be decoded and displayed in real-time. In order to achieve the required performance, we offload the decoding to the GPU using Nvidia NVDEC, which on our system with a GeForce GTX 1070 GPU was able to decode up to 14 such streams in real-time.

With the decoding being done on the GPU, and the video frames residing in GPU memory after decoding, rendering in real-time and at full resolution is easily achieved. The frames are only moved into RAM, a relatively slow operation, a few times per second in order to calculate the white balance and exposure correction on the CPU, and when the object recognition-module requests a new frame.

With the object tracking written as a separate Python application, ZeroMQ \cite{zeromq} is used for inter-process communication. The object detection also runs on a GPU, and since it is essential not to degrade the performance of the live video view, a separate GPU (GeForce GTX 1080 Ti) is used for this task. This also has the benefit that if the object tracking code were to experience a crash or a slow-down, it will not inhibit the operator. She will still get a live video stream while the object recognition module recovers. For a full block diagram of the application, see Figure ~\ref{fig:block_diagram}.

\section{Results}

\subsection{Qualitative}

\subsubsection*{Expert user feedback}
A questionnaire answered by four ATCOs provided qualitative expert user feedback on a preliminary version of the developed functionality. The first three sections of the questionnaire contained a mixture of rating-scale questions and open-ended free text questions.  The rating-scale questions were based on a four-point Likert-type scale regarding the usefulness of a particular feature, with answers ranging from 0 (not useful) to 3 (very useful). The final section provided space for the respondents to provide any additional comments that they felt were not covered by the questions.

The general consensus was that the developed technologies are promising, but require more testing. We summarize the feedback here by functionality (cf. Figure \ref{fig:Map}):
\begin{itemize}
    \item \emph{Tracking}: Overall, the ATCOs considered the tracking with 3D integration as either somewhat useful or useful. At the time of the questionnaire, unsteady bounding boxes were mentioned as distracting. In a later version this was largely resolved, leaving only a minor wobbling. The object classification abstraction level was determined to be sufficient, and the possibility of adding a class for foreign flying objects (e.g. birds, drones) was mentioned. It was remarked that it is harder to see smaller objects in the video stream than from an air traffic control tower, and that visual tracking could help to increase their visibility. 
    \item \emph{3D event positioning}: It was mentioned that interaction with the 3D model improved depth perception and situational awareness, and the various 3D functionalities were considered either somewhat useful or useful. Most ATCOs agreed that more testing is needed regarding the situational awareness and reliability.
    \item \emph{Map view}: Overall, the map view was considered either somewhat useful or useful. Some of the ATCOs found the shaded overlay useful, and some did not. In the map view, the orthophoto view was generally preferred over the abstract map view, as it is easier to relate to the video streams.
\end{itemize}


\subsubsection*{Exposure correction}
The exposure correction predominantly yields a mosaic with natural transitions, as visualized for cloudy and sunny weather conditions in Figure~\ref{fig:cams-original-corrected}. This is also the case for video, in the sense that also temporal changes generally seem natural.

In the presence of moving objects, the method generates natural results most of the time. However, the method can struggle when moving objects cross the image seams, sometimes resulting in a local flickering. Typically, the problem is most pronounced right before and after a full crossing of the seam, i.e., when the object is fully present in the boundary band of one of the images but not in the other.

Table \ref{tab:flickering} shows the results of the proposed exposure correction methods with 16 and 64 blocks vertically, when applied to concatenated video streams with a moving object right after a full crossing of the seam. The original concatenated image is shown twice for easy comparison with the correction methods.

The standard exposure correction introduces a noticeable discolouration in the block next to the car, both for large and small blocks. The object removal approach shows natural results if the remaining number of pixels in the block is relatively high (left case). However, if the moving object fills most of the block (right case), too few pixels remain for computing a natural exposure correction. The exponential smoothing approach generally shows natural results. It does, however, add a slight delay to the update of the exposure correction. For this reason we prefer using the object removal approach when applicable.

\begin{table}[h!]
\centering
\noindent\begin{tabular*}{0.6275\columnwidth}{@{\extracolsep{\stretch{1}}}*{2}{c}}
\toprule
16 blocks & 64 blocks\\
\midrule
\multicolumn{2}{c}{Original frame (duplicate)}\vspace{0.25em} \\
\includegraphics[width=0.3\columnwidth, clip, trim=0 15 0 15]{orig_frame_46_crop} &
\includegraphics[width=0.3\columnwidth, clip, trim=0 15 0 15]{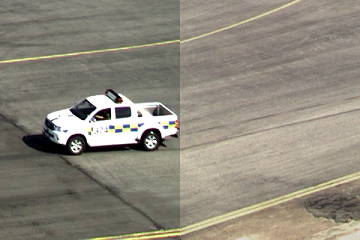} \\ \midrule
\multicolumn{2}{c}{Standard exposure correction}\vspace{0.25em} \\
\includegraphics[width=0.3\columnwidth, clip, trim=0 15 0 15]{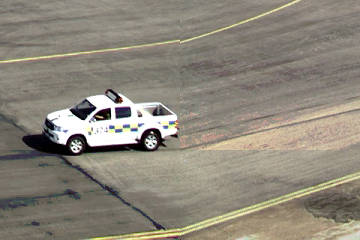} &
\includegraphics[width=0.3\columnwidth, clip, trim=0 15 0 15]{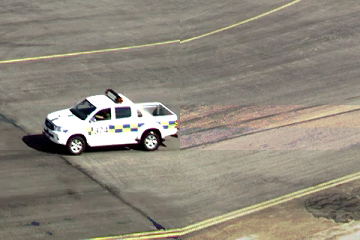}\vspace{0.25em} \\ \midrule
\multicolumn{2}{c}{Exposure correction with object removal} \\
\includegraphics[width=0.3\columnwidth, clip, trim=0 15 0 15]{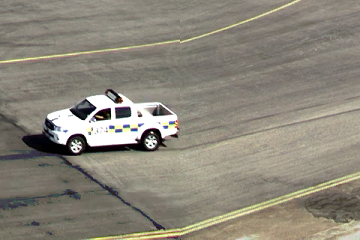} & \includegraphics[width=0.3\columnwidth, clip, trim=0 15 0 15]{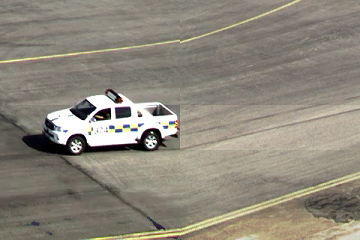}\\ \midrule
\multicolumn{2}{c}{Exposure correction with exponential smoothing}\vspace{0.25em} \\
\includegraphics[width=0.3\columnwidth, clip, trim=0 15 0 15]{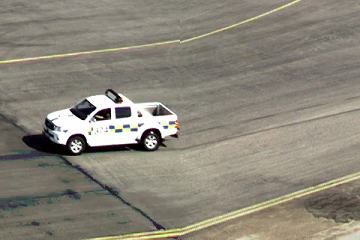} &
\includegraphics[width=0.3\columnwidth, clip, trim=0 15 0 15]{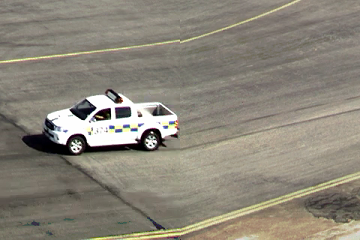}\\
\bottomrule
\end{tabular*}
\caption{Exposure correction methods with various block sizes}\label{tab:flickering}
\end{table}

\subsubsection*{Camera calibration}
Manually tuning several parameters (position, view direction, field of view, etc.) for aligning the 3D model to the video streams is a demanding process. It typically involves a field trip, expensive measuring equipment, and it can take several person-days for obtaining an accurate result.

On the other hand, the camera calibration application we developed provides a virtual environment in which the calibration can be performed, requiring only video/image data and a DSM (GeoTIFF).
The application thereby greatly sped this process up to the order of minutes.

\subsection{Quantitative}
The methods were run on a single workstation. Efficiently exploiting hardware resources and reducing unnecessary computations makes it possible to achieve real-time performance on consumer-grade hardware.

\subsubsection*{Exposure correction}
Running the exposure correction on 14 HD cameras at 30 FPS simultaneously introduced only a minor overhead on the GPU (GTX 1070). 
While visual inspection can indicate improvement of the quality of the seam, it is a highly subjective metric and difficult to judge consistently for a long scenario with up to 14 cameras. For a quantitative evaluation of the performance of our exposure correction algorithm, we use a cost function based on the method described in \cite{levin:2004}. 

Consider adjacent images $I^-$ (left) and $I^+$ (right) of size $M\times N$ with columns $I^-_{-1}, I^-_0, I^+_0, I^+_1$ from left to right of the seam. A naive measure of continuity is to directly compare the columns at the seam, i.e., $\frac1N \sum_{i=1}^N \|I^-_{0,i} - I^+_{0,i}\|$, but this measure is sensitive to geometric misalignment and asymmetrical details. Instead, the trend at row $i$ can be captured by measuring whether the gradients between the final (resp. initial) image column $(r,g,b)$ pixels continues across the seam, i.e., whether 
\[d^{\pm}_i := \|(I^\pm_{\pm 1,i} - I^\pm_{0,i}) - (I^\pm_{0,i} - I^\mp_{0,i})\| \approx 0.\]
Hence the total discrepancy of the trend can be measured as
\begin{equation}\label{eq:gradientcost}
\frac1N \sum_{i=1}^N \frac{d^+_i + d^-_i}{2}.
\end{equation}
To further reduce the contribution from geometric misalignment we down-sampled the input frames by a factor of eight in both directions.

\begin{table}[h!]
\centering
\noindent\begin{tabular*}{0.875\columnwidth}{@{\extracolsep{\stretch{1}}}*{3}{c}}
\toprule
image interior & image seam & image seam\\ 
               & uncorrected & corrected\\ 
\midrule
\multicolumn{3}{c}{sunny~\quad}\vspace{0.25em}\\
\includegraphics[width=0.275\columnwidth, clip, trim=135 30 135 30]{sunny-uncorrected} &
\includegraphics[width=0.275\columnwidth, clip, trim=202.5 30 67.5 30]{sunny-uncorrected} & 
\includegraphics[width=0.275\columnwidth, clip, trim=202.5 30 67.5 30]{sunny-corrected}\vspace{0.25em}\\
8.62 & 49.82 & 19.75 \vspace{0.25em}\\ \midrule
\multicolumn{3}{c}{cloudy~\quad}\vspace{0.25em}\\
\includegraphics[width=0.275\columnwidth, clip, trim=270 60 270 60]{cloudy-uncorrected} &
\includegraphics[width=0.275\columnwidth, clip, trim=405 60 135 60]{cloudy-uncorrected} & 
\includegraphics[width=0.275\columnwidth, clip, trim=405 60 135 60]{cloudy-corrected}\vspace{0.25em}\\
6.37 & 61.19 & 16.11 \vspace{0.25em}\\
\bottomrule
\end{tabular*}
\caption{Average values of the cost function \eqref{eq:gradientcost}}\label{tab:gradientcost} 
\end{table}

Table \ref{tab:gradientcost} shows the average value of \eqref{eq:gradientcost} for two scenes with 6 cameras and a duration of 60 seconds. The reference value in the left columns was evaluated at the middle of the input streams. The exposure corrected result is a significant improvement over the initial uncorrected seam, but is still significantly higher than the reference value. This deviation can partly be explained by a slight geometric misalignment at the seam.

\section{Conclusion}
In this paper, we have developed and tested a number of techniques based on video processing, 3D modelling and object tracking that apply to high resolution video arising from remote towers.

It was shown that the methods can be implemented on a single workstation and still retain real-time performance by efficiently exploiting hardware resources and by reducing unnecessary computations. The techniques do not rely on expensive or special-made hardware, thereby supporting the cost-effectiveness of the remote tower concept by limiting start-up and maintenance costs.

Results from a questionnaire answered by ATCOs indicated that the developed technologies are promising, but require more testing. The visual tracking was remarked to have the potential to increase visibility, both of small objects and using night-vision technologies, with the potential to improve safety.

As future work, there are several possibilities for improving the proposed functionality. The attention mechanisms considered in this paper are based on detecting movements and expected locations of existing tracked objects. In the future, we could also consider where ATCOs concentrate their attention, by looking at heat maps from tracked eye movements \cite{li2018much} in order to attain better performance. 

The tracking functionality has so far only been tested in relatively high visibility conditions during daytime. More testing is needed to see how reliable the tracking is in low visibility and nighttime scenarios. Testing using infrared sensors is also subject to future work.

Although the exposure correction works well in general, there are still situations where it could be improved. The exposure correction maps currently act on each colour channel separately. The quality of the corrections can be expected to improve when using linear maps combining the three channels, at a negligible computational overhead. Moreover, currently the exposure correction map is defined separately for each vertical block. The transition between these maps could be improved by either using convex combinations of the adjacent maps or by adding boundary conditions. Finally, more tuning is needed for automatically selecting which of the proposed methods to use.


\section*{Acknowledgments}
This project has received funding from the SESAR Joint Undertaking under the European Union's Horizon 2020 research and innovation programme under grant agreement No 730195.

The authors would like to thank Saab and Luftfartsverket Sweden (LFV) for their assistance in providing access to the remote tower video data from Sundsvall, Sweden. We also express our gratitude to the four ATCOs from the COOPANS partnership for their valuable feedback.

\bibliography{main}

\end{document}